\documentclass[11pt,a4paper]{article}
\pdfoutput=1
\usepackage[hyperref]{acl2019}
\usepackage{times}
\usepackage{microtype}

\usepackage{url}

\usepackage{amsmath}
\usepackage{amsfonts}
\usepackage{amssymb}

\usepackage{mathptmx}	
\usepackage[scaled=.8]{beramono}
\usepackage[scaled=.85]{helvet}
\usepackage[T1]{fontenc}
\usepackage[utf8x]{inputenc}

\usepackage{latexsym}

\usepackage{titlesec}
\titleformat*{\subparagraph}{\itshape\bfseries}
\titlespacing{\subparagraph}{%
  1em}{
  0pt}{
  1em}

\usepackage{graphicx}

\usepackage[shortlabels]{enumitem} 
\setitemize{noitemsep,topsep=0em} 
\setenumerate{noitemsep,leftmargin=0em,itemindent=13pt,topsep=0em}

\usepackage[olditem]{paralist}

\usepackage{subcaption}
\usepackage{cleveref}
\usepackage{multirow}

\crefformat{part}{\S#2#1#3}
\crefformat{chapter}{\S#2#1#3}
\crefformat{section}{\S#2#1#3}
\crefformat{subsection}{\S#2#1#3}
\crefformat{subsubsection}{\S#2#1#3}
\crefformat{paragraph}{\P#2#1#3}
\crefformat{subparagraph}{\P#2#1#3}
\crefmultiformat{section}{\S#2#1#3}{ and~\S#2#1#3}{, \S#2#1#3}{, and~\S#2#1#3}
\crefmultiformat{subsection}{\S#2#1#3}{ and~\S#2#1#3}{, \S#2#1#3}{, and~\S#2#1#3}
\crefmultiformat{subsubsection}{\S#2#1#3}{ and~\S#2#1#3}{, \S#2#1#3}{, and~\S#2#1#3}
\crefmultiformat{paragraph}{\P\P#2#1#3}{ and~#2#1#3}{, #2#1#3}{, and~#2#1#3}
\crefmultiformat{subparagraph}{\P\P#2#1#3}{ and~#2#1#3}{, #2#1#3}{, and~#2#1#3}
\crefrangeformat{section}{\mbox{\S\S#3#1#4--#5#2#6}}
\crefrangeformat{subsection}{\mbox{\S\S#3#1#4--#5#2#6}}
\crefrangeformat{subsubsection}{\mbox{\S\S#3#1#4--#5#2#6}}
\crefrangeformat{paragraph}{\mbox{\P\P#3#1#4--#5#2#6}}
\crefrangeformat{subparagraph}{\mbox{\P\P#3#1#4--#5#2#6}}
\crefname{part}{Part}{Parts}
\Crefname{part}{Part}{Parts}
\crefname{chapter}{ch.}{ch.}
\Crefname{chapter}{Ch.}{Ch.}
\crefname{figure}{figure}{figures}
\crefname{subfigure}{figure}{figures}
\Crefname{subfigure}{Figure}{Figures}
\crefname{appsec}{appendix}{appendices}
\Crefname{appsec}{Appendix}{Appendices}
\crefname{algocf}{algorithm}{algorithms}
\Crefname{algocf}{Algorithm}{Algorithms}
\crefname{enums,enumsi}{example}{examples}
\Crefname{enums,enumsi}{Example}{Examples}
\crefname{}{example}{examples} 
\Crefname{}{Example}{Examples}
\crefformat{enums}{(#2#1#3)}
\crefformat{enumsi}{(#2#1#3)}
\crefrangeformat{enums}{\mbox{(#3#1#4--#5#2#6)}}
\crefrangeformat{enumsi}{\mbox{(#3#1#4--#5#2#6)}}
\crefformat{}{(#2#1#3)}
\crefname{xnumi}{example}{examples} 
\crefname{xnumi}{example}{examples} 
\Crefname{xnumii}{Example}{Examples} 
\Crefname{xnumii}{Example}{Examples} 
\crefformat{xnumi}{(#2#1#3)} 
\crefformat{xnumii}{(#2#1#3)} 
\crefrangeformat{enums}{\mbox{(#3#1#4--#5#2#6)}}
\crefrangeformat{enumsi}{\mbox{(#3#1#4--#5#2#6)}}
\crefrangeformat{xnumi}{\mbox{(#3#1#4--#5#2#6)}} 
\crefrangeformat{xnumii}{\mbox{(#3#1#4--#5#2#6)}} 
\crefmultiformat{enumsi}{(#2#1#3}{, #2#1#3)}{, #2#1#3}{, #2#1#3)}
\crefmultiformat{xnumi}{(#2#1#3}{, #2#1#3)}{, #2#1#3}{, #2#1#3)} 
\crefmultiformat{xnumii}{(#2#1#3}{, #2#1#3)}{, #2#1#3}{, #2#1#3)} 
\crefrangemultiformat{enumsi}{(#3#1#4--#5#2#6}{, #3#1#4--#5#2#6)}{, #3#1#4--#5#2#6}{, #3#1#4--#5#2#6)}
\crefrangemultiformat{xnumi}{(#3#1#4--#5#2#6}{, #3#1#4--#5#2#6)}{, #3#1#4--#5#2#6}{, #3#1#4--#5#2#6)} 
\crefrangemultiformat{xnumii}{(#3#1#4--#5#2#6}{, #3#1#4--#5#2#6)}{, #3#1#4--#5#2#6}{, #3#1#4--#5#2#6)} 

\ifx\creflastconjunction\undefined%
\newcommand{\creflastconjunction}{, and\nobreakspace} 
\else%
\renewcommand{\creflastconjunction}{, and\nobreakspace} 
\fi%

\newcommand*{\Fullref}[1]{\hyperref[{#1}]{\Cref*{#1}: \nameref*{#1}}}
\newcommand*{\fullref}[1]{\hyperref[{#1}]{\cref*{#1}: \nameref{#1}}}

\usepackage{xcolor}
\usepackage{tikz}
\usepackage{tikz-dependency} 
\usepackage{forest}

\DeclareMathOperator*{\argmax}{arg\,max}

\usepackage[plain]{algorithm}
\usepackage{algpseudocode}

\usepackage{footnote}
\usepackage[symbol]{footmisc}
\newcommand{\astfootnotetext}[1]{
\renewcommand{\thefootnote}{\fnsymbol{footnote}}
}

\newcommand{\ucca}[1]{\textcolor{gray}{\textbf{\textsf{#1}}}}

\algnewcommand\algorithmicforeach{\textbf{for each}}
\algdef{S}[FOR]{ForEach}[1]{\algorithmicforeach\ #1\ \algorithmicdo}

\useforestlibrary{linguistics}
\forestapplylibrarydefaults{linguistics}

\makeatletter
\renewcommand{\paragraph}{%
  \@startsection{paragraph}{4}%
  {\z@}{.2ex \@plus 1ex \@minus .2ex}{-1em}%
  {\normalfont\normalsize\bfseries}%
}
\makeatother

\usepackage{color}
\usepackage{bm}
\definecolor{orange}{rgb}{1,0.5,0}
\definecolor{mdgreen}{rgb}{0.05,0.6,0.05}
\definecolor{mdblue}{rgb}{0,0,0.7}
\definecolor{dkblue}{rgb}{0,0,0.5}
\definecolor{dkgray}{rgb}{0.3,0.3,0.3}
\definecolor{slate}{rgb}{0.25,0.25,0.4}
\definecolor{gray}{rgb}{0.5,0.5,0.5}
\definecolor{ltgray}{rgb}{0.7,0.7,0.7}
\definecolor{purple}{rgb}{0.7,0,1.0}
\definecolor{lavender}{rgb}{0.65,0.55,1.0}

\newcommand{\w}[1]{\textit{#1}}  

\newcommand{\finalversion}[1]{}
\newcommand{\shortversion}[1]{}

\newcommand{\draftnotice}[1]{} 
\newcommand{\anonversion}[1]{} 
\newcommand{\nonanonversion}[1]{#1} 

\newcommand{\ensuretext}[1]{#1}
\newcommand{\nssmarker}{\ensuretext{\textcolor{magenta}{\ensuremath{^{\textsc{NS}}_{\textsc{S}}}}}}

\newcommand{\arkcomment}[3]{\ensuretext{\textcolor{#3}{[#1 #2]}}}
\newcommand{\nss}[1]{\arkcomment{\nssmarker}{#1}{magenta}}

\aclfinalcopy 
\def\aclpaperid{22} 

\title{Semantically Constrained Multilayer Annotation:\\ The Case of Coreference}

\nonanonversion{
\author{
  Jakob Prange\footnotemark{} \quad Nathan Schneider \\
  Georgetown University \\
  \And
  Omri Abend \\
  The Hebrew University of Jerusalem \\
}
}

\date{}

\begin{document}
\maketitle

\footnotetext[1]{Contact: \url{jakob@cs.georgetown.edu}}
\renewcommand{\thefootnote}{\arabic{footnote}}

\begin{abstract}
We propose a coreference annotation scheme as a layer on top of 
the Universal Conceptual Cognitive Annotation foundational layer, treating units in predicate-argument structure as a basis for entity and event mentions.
We argue that this allows coreference annotators to sidestep 
some of the challenges faced in other schemes, 
which do not enforce consistency with predicate-argument structure and vary widely in what kinds of mentions they annotate and how. 
The proposed approach is examined with a pilot annotation study and compared with annotations from other schemes.
\end{abstract}

\section{Introduction}

Unlike some NLP tasks, coreference resolution lacks an agreed-upon standard for annotation and evaluation \citep{poesio-16}. 
It has been approached using a multitude of different markup schemas, and the several evaluation metrics commonly used \citep{pradhan-14} are controversial \citep{moosavi-16}.
In particular, these schemas use divergent and often (language-specific) syntactic criteria for defining candidate mentions in text. 
This includes the questions of whether to annotate entity and/or event coreference, whether to include singletons, and how to identify the precise span of complex mentions.
Recognition of this limitation in the field has recently prompted the Universal Coreference initiative,\footnote{\url{https://sites.google.com/view/crac2019/}} which aims to settle on a single cross-linguistically applicable annotation standard.

We think that many issues stem from the common practice of creating mention annotations from scratch on the raw or tokenized text, and we suggest that they could be overcome by reusing structures from existing semantic annotation, thereby ensuring compatibility between the layers. 
We advocate for the design pattern of a \textbf{semantic foundational layer}, which defines a basic semantic structure that additional layers can refine or make reference to.
Some form of predicate-argument structure involving entities and propositions should serve as a natural semantic foundation for a layer that groups coreferring entity and event mentions into clusters.

Here we argue that Universal Conceptual Cognitive Annotation \citep[UCCA;][]{abend-13} is an ideal choice, as it defines a foundational layer of predicate-argument structure whose main design principles are cross-linguistic applicability and fast annotatability by non-experts.
To that end, we develop and pilot a new layer for UCCA which adds coreference information.%
\footnote{Our annotations will be made available under \\\url{https://github.com/jakpra/UCoref}.}
This coreference layer is constrained by the spans already specified in the foundational predicate-argument layer.
We compare these manual annotations to existing gold coreference annotations in multiple frameworks\finalversion{\nss{Maybe:}, and to system predictions}, 
finding a healthy level of overlap.

Our contributions are:
\begin{itemize}
\item A discussion of multilayer design principles informed by existing semantically annotated corpora (\cref{sec:background}).
\item A semantically-based framework for mention identification and coreference resolution as a layer of UCCA  (\cref{sec:desc}). Reusing UCCA units as mentions facilitates efficient and consistent multilayer annotation.
We call the framework \textbf{Universal Conceptual Cognitive Coreference} (UCoref).
\item An in-depth comparison to three other coreference frameworks based on annotation guidelines (\cref{sec:compare}) and a pilot English dataset  (\cref{sec:pilot}).
\end{itemize}

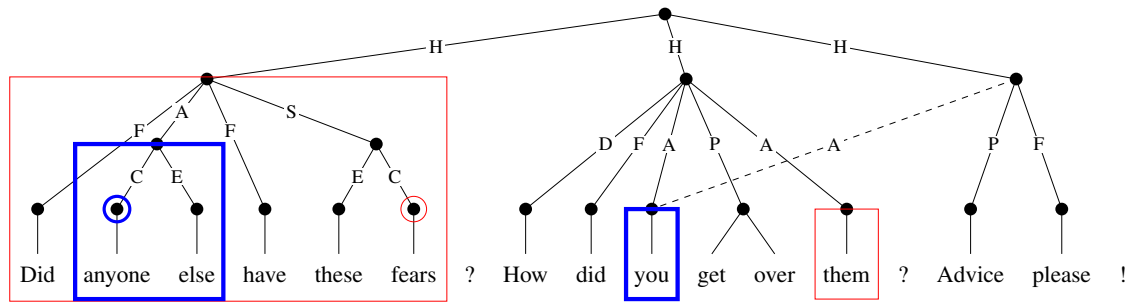
\begin{figure*}[ht]
\centering
\scalebox{.8}{
\begin{forest}
for tree={parent anchor=north, child anchor=north, s sep=0.5em,l sep=.5em}
[
    [, edge label={node[midway,fill=white,inner sep=1pt,font=\footnotesize]{H}}, tikz={\node [draw,red,inner sep=1,fit to=tree]{};},
      [, edge label={node[pos=0.4,fill=white,inner sep=1pt,font=\footnotesize]{F}}, tier=word [Did]]
      [, edge label={node[midway,fill=white,inner sep=1pt,font=\footnotesize]{A}}, tikz={\node [draw,blue,inner sep=0,fit to=tree,line width=2pt]{};},
        [, edge label={node[midway,fill=white,inner sep=1pt,font=\footnotesize]{C}}, tier=word [anyone]]
        [, edge label={node[midway,fill=white,inner sep=1pt,font=\footnotesize]{E}}, tier=word [else]]
      ]
      [, edge label={node[pos=0.4,fill=white,inner sep=1pt,font=\footnotesize]{F}}, tier=word [have]]
      [, edge label={node[midway,fill=white,inner sep=1pt,font=\footnotesize]{S}},
        [, edge label={node[midway,fill=white,inner sep=1pt,font=\footnotesize]{E}}, tier=word [these]]
        [, edge label={node[midway,fill=white,inner sep=1pt,font=\footnotesize]{C}}, tier=word [fears, tier=tok]]
      ]
    ]
    [?, no edge, tier=tok]
    [, edge label={node[midway,fill=white,inner sep=1pt,font=\footnotesize]{H}},
      [, edge label={node[midway,fill=white,inner sep=1pt,font=\footnotesize]{D}}, tier=word [How]]
      [, edge label={node[midway,fill=white,inner sep=1pt,font=\footnotesize]{F}}, tier=word [did]]
      [, edge label={node[midway,fill=white,inner sep=1pt,font=\footnotesize]{A}}, tikz={\node [draw,blue,inner sep=0,fit to=tree,line width=2pt]{};}, tier=word [you]]
      [, edge label={node[midway,fill=white,inner sep=1pt,font=\footnotesize]{P}}, tier=word
        [get]
        [over]
      ]
      [, edge label={node[midway,fill=white,inner sep=1pt,font=\footnotesize]{A}}, tikz={\node [draw,red,inner sep=0,fit to=tree]{};}, tier=word [them]]
    ]
    [?, no edge, tier=tok]
    [, edge label={node[midway,fill=white,inner sep=1pt,font=\footnotesize]{H}},
      [, edge label={node[midway,fill=white,inner sep=1pt,font=\footnotesize]{P}}, tier=word [Advice]]
      [, edge label={node[midway,fill=white,inner sep=1pt,font=\footnotesize]{F}}, tier=word [please]]
    ]
    [!, no edge, tier=tok]
]
\path[fill=black] (.parent anchor) circle[radius=3pt] 
                 (!1.child anchor) circle[radius=3pt]
                 (!11.child anchor) circle[radius=3pt]
                 (!12.child anchor) circle[radius=3pt]
                 (!121.child anchor) circle[radius=3pt]
                 (!122.child anchor) circle[radius=3pt]
                 (!13.child anchor) circle[radius=3pt]
                 (!14.child anchor) circle[radius=3pt]
                 (!141.child anchor) circle[radius=3pt]
                 (!142.child anchor) circle[radius=3pt]
                 (!3.child anchor) circle[radius=3pt]
                 (!31.child anchor) circle[radius=3pt]
                 (!32.child anchor) circle[radius=3pt]
                 (!33.child anchor) circle[radius=3pt]
                 (!34.child anchor) circle[radius=3pt]
                 (!35.child anchor) circle[radius=3pt]
                 (!5.child anchor) circle[radius=3pt]
                 (!51.child anchor) circle[radius=3pt]
                 (!52.child anchor) circle[radius=3pt];
\path[draw=blue,line width=1.5pt] (!121.child anchor) circle[radius=6pt];
\path[draw=red] (!142.child anchor) circle[radius=6pt];
\draw[dashed] (!5.child anchor) to node(you-remote)[midway,fill=white,inner sep=1pt,font=\footnotesize]{A} (!33.child anchor);
\end{forest}
}
\caption{\label{fig:intro-ex}A foundational UCCA analysis of three consecutive sentences from the Richer Event Description corpus, with examples of coreferent units superimposed (boxes).
The context is that the speaker is posting a message to a forum in which she shares her own fears and asks for advice; \textit{you} is coreferent with \textit{anyone else}, and \textit{them} refers back to the whole first scene.\footnotemark{} Circled nodes indicate semantic heads\slash minimal spans, as determined by following State (\ucca{S}) and Center (\ucca{C}) edges. In the third sentence, \textit{Advice please!}, the addressee/adviser is a salient, but implicit Participant (\ucca{A}) which is expressed with a remote (dashed) edge to a prior mention.
Remaining categories are abbreviated as: \ucca{H} -- Parallel Scene, \ucca{P} -- Process, \ucca{E} -- Elaborator, \ucca{D} -- Adverbial, \ucca{F} -- Function.
}
\label{fig:threesents}
\end{figure*}

\section{Background and Motivation}\label{sec:background}

We first consider the organization of semantic annotations in corpora, arguing that UCCA's representation of predicate-argument structure should serve as a foundation for coreference annotations.

\footnotetext{Following UCCA's philosophy, we interpret both \textit{fears} and \textit{them} mainly as evoking the emotional state of having fears (i.e., ``how did you get over them'' $\approx$ ``how did you get over being afraid'').
This analysis abstracts away from the more direct reading as the specific objects of fear; but either way, the proper semantic head of the first sentence has to be \textit{fears} (not \textit{have}), and from our flexible minimum\slash maximum span policy it follows that any mention coreferring with \textit{fears} automatically corefers with the whole scene.

Further, we interpret both \textit{anyone else} and \textit{you} as referring to the unknown-sized set of audience members sharing the speaker's fears.
Whereas \textit{you} introduces a presupposition that this set is non-empty, this is not the case for the negative polarity item \textit{anyone else}.
Although questionable in terms of cohesion (as the presupposition created by \textit{you} fails if the answer to the first question is `no'), this is a typical phenomenon in conversational data and can be explained by recognizing that the second question is implicitly conditional: ``\textbf{If so,} how did you get over them?''}

\subsection{Approaches to Semantic Multilayering}\label{sec:pred-arg}

A major consideration in the design of coreference annotation schemes, as well as meaning representations generally, is what the relevant annotation targets are and whether they should be normalized across layers when the text is annotated for multiple aspects of linguistic structure.
Should coreference be annotated completely independently of decisions about syntactic phrases and semantic predicate-argument structures?
On the one hand, this decoupling of the annotations might absolve the coreference annotators from having to worry about other annotation conventions in the corpus. 
On the other hand, this is potentially a recipe for inconsistent annotations across layers, making it more difficult to integrate information across layers for complex reasoning in natural language understanding systems. 
Moreover, certain details of coreference annotation may be underdetermined such that relying on other layers would save coreference annotators and guidelines-developers from having to reinvent the wheel.

We can examine existing semantic annotation schemes with regard to two closely related criteria: a)~\textbf{anchoring}, i.e.~the previously determined underlying structure (characters, tokens, syntax, etc.)\ that defines the set of possible annotation targets in a new layer; and b)~\textbf{modularity}, the extent to which multiple kinds of information are expressed as separate (possibly linked) structures\slash layers, which may be annotated in different phases.

\paragraph*{Massively multilayer corpora.}
A few corpora comprise several layers of annotation, including semantics, with an emphasis on modularity of these layers. 
One example is OntoNotes \citep{ontonotes}, 
annotated for syntax, named entities, word senses, PropBank \citep{palmer-05} predicate-argument structures, and coreference.
Another example is GUM \citep{zeldes-17}, with layers for syntactic, coreference, discourse, and document structure. 
Both of these resources cover multiple genres.
Different layers in these resources are anchored differently, as noted below.

\paragraph*{Token-anchored.}
Many semantic annotation layers are specified in terms of character or token offsets. This is the case for UCCA's Foundational Layer (\cref{sec:ucca}), FrameNet \citep{framenet}, RED \citep{ogorman-16}, all of the layers in GUM, and the named entity and word sense annotations in OntoNotes.
Though the guidelines may mention syntactic criteria for deciding what units to semantically annotate, the annotated data does not explicitly tie these layers to syntactic units, and to the best of our knowledge the annotator is not constrained by the syntactic annotation.

\paragraph*{Syntax-anchored.}
Semantic annotations explicitly defined in terms of syntactic units include: PropBank (such as in OntoNotes);
and the coreference annotations in the Prague Dependency Treebank \citep[PDT;][]{nedoluzhko-16}.
In addition, PDT's ``deep syntactic'' tectogrammatical layer, which is built on the syntactic analytic layer, can be considered quasi-semantic \citep{bohmova-03}.

\subparagraph*{Transformed syntax.}
In other cases, semantic label annotations enrich skeletal semantic representations that have been deterministically converted from syntactic structures.
One example is Universal Decompositional 
Semantics \citep{white-16}, whose annotations
are anchored with PredPatt, a way of converting Universal Dependencies trees \citep{nivre-16} to approximate predicate-argument structures.

\paragraph*{Sentence-anchored.}
The Abstract Meaning Representation \citep[AMR;][]{amr} is an example of a highly integrative (anti-modular) approach to sentence-level meaning, without anchoring below the sentence level.
AMR annotations take the form of a single graph per sentence, capturing a variety of kinds of information, including predicate-argument structure, sentence focus, modality, lexical semantic distinctions, coreference, named entity typing, and entity linking (``Wikification'').
English AMR annotators provide the full graph at once (with the exception of entity linking, done as a separate pass), and do not mark how pieces of the graph are anchored in tokens, which has spawned a line of research on various forms of token-level alignment for parsing \citep[e.g.][]{flanigan-14,pourdamghani-14,chen-17,szubert-18,liu-18}. 
Chinese AMR, by contrast, is annotated in a way that aligns nodes with tokens \citep{li-16}.

\paragraph*{Semantics-anchored.}
The approach we explore here is the use of a \emph{semantic} layer as a foundation for a different type of semantic layer. Such approaches support modularity, while still allowing annotation reuse. A recent example for this approach is multi-sentence AMR \citep{ogorman-18}, which links together the previously annotated per-sentence AMR graphs to indicate coreference across sentences.


\subsection{UCCA's Foundational Layer}\label{sec:ucca}

UCCA is a coarse-grained, typologically-motivated scheme for analyzing abstract semantic structures in text. 
It is designed to expose commonalities in semantic structure 
across paraphrases and translations, 
with a focus on predicate-argument and other semantic head-modifier relations.
Formally, each text passage is annotated with a directed acyclic graph (DAG)
over semantic elements called \textbf{units}.
Each unit, corresponding to (anchored by) one or more tokens,
is labeled with one or more semantic \textbf{categories} in relation to a parent unit.

\begin{figure*}[ht]
\centering
\scalebox{.8}{
\begin{forest}
for tree={parent anchor=north, child anchor=north, s sep=0.5em,l sep=.5em}
[
    [, edge=lightgray, edge label={node[midway,fill=white,inner sep=1pt,font=\footnotesize]{H}}, tikz={\node [draw,red,inner sep=2,fit to=tree]{};},
      [, edge=lightgray, edge label={node[pos=0.4,fill=white,inner sep=1pt,font=\footnotesize]{F}}, tier=word [Did, edge=lightgray]]
      [, edge=lightgray, edge label={node[midway,fill=white,inner sep=1pt,font=\footnotesize]{A}}, tikz={\node [draw,blue,inner sep=0,fit to=tree]{};},
        [, edge=lightgray, edge label={node[midway,fill=white,inner sep=1pt,font=\footnotesize]{C}}, tier=word [anyone, edge=lightgray]]
        [, edge=lightgray, edge label={node[midway,fill=white,inner sep=1pt,font=\footnotesize]{E}}, tier=word [else, edge=lightgray]]
      ]
      [, edge=lightgray, edge label={node[pos=0.4,fill=white,inner sep=1pt,font=\footnotesize]{F}}, tier=word [have, edge=lightgray]]
      [, edge=lightgray, edge label={node[midway,fill=white,inner sep=1pt,font=\footnotesize]{S}},
        [, edge=lightgray, edge label={node[midway,fill=white,inner sep=1pt,font=\footnotesize]{E}}, tier=word [these, edge=lightgray]]
        [, edge=lightgray, edge label={node[midway,fill=white,inner sep=1pt,font=\footnotesize]{C}}, tier=word [fears, edge=lightgray, tier=tok]]
      ]
    ]
    [?, no edge, tier=tok]
    [, edge=lightgray, edge label={node[midway,fill=white,inner sep=1pt,font=\footnotesize]{H}}, tikz={\node [draw,purple,inner sep=2,fit to=tree]{};},
      [, edge=lightgray, edge label={node[midway,fill=white,inner sep=1pt,font=\footnotesize]{D}}, tier=word [How, edge=lightgray]]
      [, edge=lightgray, edge label={node[midway,fill=white,inner sep=1pt,font=\footnotesize]{F}}, tier=word [did, edge=lightgray]]
      [, edge=lightgray, edge label={node[midway,fill=white,inner sep=1pt,font=\footnotesize]{A}}, tikz={\node [draw,blue,inner sep=0,fit to=tree]{};}, tier=word [you, edge=lightgray]]
      [, edge=lightgray, edge label={node[midway,fill=white,inner sep=1pt,font=\footnotesize]{P}}, tier=word
        [get, edge=lightgray]
        [over, edge=lightgray]
      ]
      [, edge=lightgray, edge label={node[midway,fill=white,inner sep=1pt,font=\footnotesize]{A}}, tikz={\node [draw,red,inner sep=0,fit to=tree]{};}, tier=word [them, edge=lightgray]]
    ]
    [?, no edge, tier=tok]
    [, edge=lightgray, edge label={node[midway,fill=white,inner sep=1pt,font=\footnotesize]{H}}, tikz={\node [draw,olive,inner sep=2,fit to=tree]{};},
      [, edge=lightgray, edge label={node[midway,fill=white,inner sep=1pt,font=\footnotesize]{P}}, tier=word [Advice, edge=lightgray]]
      [, edge=lightgray, edge label={node[midway,fill=white,inner sep=1pt,font=\footnotesize]{F}}, tier=word [please, edge=lightgray]]
    ]
    [!, no edge, tier=tok]
]
\path[fill=lightgray] (.parent anchor) circle[radius=3pt] 
                 (!11.child anchor) circle[radius=3pt]
                 (!121.child anchor) circle[radius=3pt]
                 (!122.child anchor) circle[radius=3pt]
                 (!13.child anchor) circle[radius=3pt]
                 (!14.child anchor) circle[radius=3pt]
                 (!141.child anchor) circle[radius=3pt]
                 (!142.child anchor) circle[radius=3pt]
                 (!3.child anchor) circle[radius=3pt]
                 (!31.child anchor) circle[radius=3pt]
                 (!32.child anchor) circle[radius=3pt]
                 (!34.child anchor) circle[radius=3pt]
                 (!35.child anchor) circle[radius=3pt]
                 (!5.child anchor) circle[radius=3pt]
                 (!51.child anchor) circle[radius=3pt]
                 (!52.child anchor) circle[radius=3pt];
\node[draw=olive,star,star points=7,inner sep=0pt,minimum size=15pt] at (!51.child anchor) {};
\node[draw=purple,regular polygon,regular polygon sides=3,inner sep=0pt,minimum size=15pt] at (!34.child anchor) {};
\node[draw=blue,diamond,inner sep=0pt,minimum size=15pt] at (!121.child anchor) {};
\node[draw=red,rectangle,inner sep=0pt,minimum size=10pt] at (!142.child anchor) {};
\node[fill=olive,star,star points=7,inner sep=0pt,minimum size=15pt] at (!5.child anchor) {};
\node[fill=purple,regular polygon,regular polygon sides=3,inner sep=0pt,minimum size=15pt] at (!3.child anchor) {};
\node[fill=blue,diamond,inner sep=0pt,minimum size=15pt] at (!12.child anchor) {};
\node[fill=blue,diamond,inner sep=0pt,minimum size=15pt] at (!33.child anchor) {};
\node[fill=red,rectangle,inner sep=0pt,minimum size=10pt] at (!1.child anchor) {};
\node[fill=red,rectangle,inner sep=0pt,minimum size=10pt] at (!35.child anchor) {};
\draw[dashed,lightgray] (!5.child anchor) to node(you-remote)[midway,fill=white,inner sep=1pt,font=\footnotesize]{A} (!33.child anchor);
(!51.child anchor);
\path let \p1 = (!141.child anchor) in node(e1)[diamond,inner sep=0pt,minimum size=15pt,fill=blue]  at (\x1,.2) {};
\path let \p1 = (e1) in node[inner sep=0pt,text=blue]  at (-6.5,\y1) {\textit{addressee}};
\draw[blue] (e1) -- (!12.child anchor);
\draw[blue] (e1) -- (!33.child anchor);
\path let \p1 = (!2.child anchor) in node(e2)[rectangle,inner sep=0pt,minimum size=10pt,fill=red]  at (\x1,.2) {};
\path let \p1 = (e2) in node[inner sep=0pt,text=blue]  at (-1.4,\y1) {\textit{addressee}};
\node[inner sep=0pt,text=red] at (-1.4,-.2) {\textit{has fears}};
\draw[red] (e2) -- (!1.child anchor);
\draw[red] (e2) -- (!35.child anchor);
\path let \p1 = (!3.child anchor) in node(e3)[regular polygon,regular polygon sides=3,inner sep=0pt,minimum size=15pt,fill=purple]  at (\x1,.2) {};
\path let \p1 = (e3) in node[inner sep=0pt,text=blue]  at (1.6,\y1) {\textit{addressee}};
\path let \p1 = (e3) in node[inner sep=0pt,text=purple]  at (1.65,-.2) {\textit{surmounts}};
\path let \p1 = (e3) in node[inner sep=0pt,text=red]  at (1.25,-.6) {\textit{fears}};
\draw[purple] (e3) -- (!3.child anchor);
\path let \p1 = (!5.child anchor) in node(e4)[star,star points=7,inner sep=0pt,minimum size=15pt,fill=olive]  at (\x1,.2) {};
\path let \p1 = (e4) in node[inner sep=0pt,text=blue]  at (4.5,\y1) {\textit{addressee}};
\node[inner sep=0pt,text=olive]  at (4.7,-.2) {\textit{advises}};
\draw[olive] (e4) -- (!5.child anchor);
\end{forest}
}
\caption{\label{fig:ref-layer}The reference layer UCoref on top of UCCA's foundational layer. %
A new ``referent node'' is introduced as a parent for each cluster of coreferring mentions.
Colors and shapes indicate coreferring mentions. 
By virtue of the remote Participant edge (dashed line), 
the addressee referent implicitly participates in the third scene as well.}
\end{figure*}
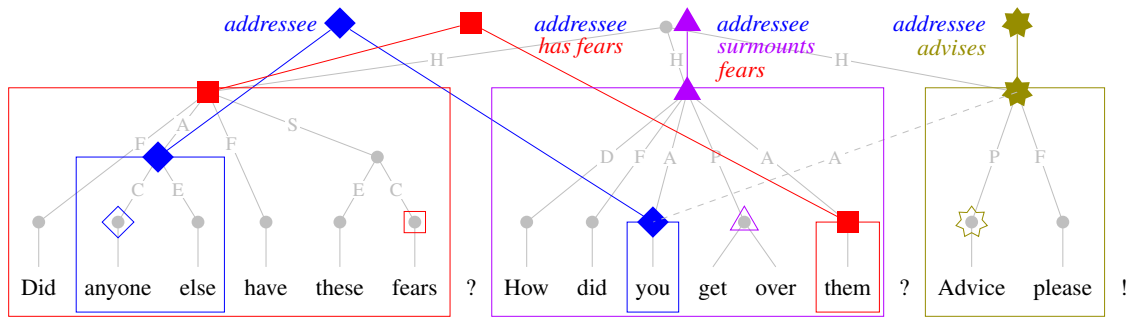

The foundational layer\footnote{Annotation guidelines: \url{https://github.com/UniversalConceptualCognitiveAnnotation/docs/blob/master/guidelines.pdf}} specifies a DAG structure organized in terms of \textbf{scenes} (events\slash situations mentioned in the text). This can be seen for three sentences in \cref{fig:threesents}, 
where each corresponds to a Parallel Scene (denoted by the category label \ucca{H}) as three events are presented in sequence. 
A scene unit is headed by a predicate, which is either a State (\ucca{S}), like \w{these fears}, or a Process (\ucca{P}), like \w{get over}. 
Most scenes have at least one Participant (\ucca{A}), 
typically an entity or location---in this case, the individuals experiencing fear.
Semantic refinements of manner, aspect, modality, negation, causativity, etc.\ are marked with the category Adverbial (\ucca{D}). 
Time (\ucca{T}) is used for temporal modifiers.
Within a non-scene unit, the semantic head is marked Center (\ucca{C}), while semantic modifiers are Elaborators (\ucca{E}). 
Function (\ucca{F}) applies to words considered to add no semantic content relevant to the scene structure.

Some additional structural properties are worthy of note. 
An \textbf{unanalyzable unit} indicates that a group of tokens form a multiword expression 
with no internal semantic structure, like \w{get over} `surmount'.
A \textbf{remote edge} (reentrancy, shown as a dashed line in \cref{fig:threesents}) makes it possible for a unit to have multiple parent units such that the structure is not a tree. 
This is mainly used when a Participant is shared by multiple scenes.
Texts are annotated in passages generally larger than sentences, 
and remote edges may cross sentence boundaries---for example, when a Participant mentioned in one sentence is implicit in the next, such as \w{you} as the implicit advice-giver in the sentence \w{Advice please!}.
\textbf{Implicit units} are null elements used when there is a salient piece of the meaning that is implied but not expressed overtly \emph{anywhere} in the passage. 
(If the third sentence from \cref{fig:threesents} was annotated in isolation, the advice-giver would be represented by an implicit unit.)


\subsection{Insufficiency of the Foundational Layer}\label{sec:ucca-motiv}

In addition to the benefits of a semantic foundational layer for coreference annotation (\cref{sec:pred-arg}), we point out how adding such a layer to UCCA would rectify shortcomings of the foundational layer.

First and foremost, UCCA currently lacks any representation of ``true'' coreference, i.e., the phenomenon that two or more \emph{explicit} units are mentions of the same entity.
Second, though remote edges are helpful for indicating that a Participant is shared between multiple scenes, this is problematic if the referent is mentioned multiple times in the passage. 
Because the information that those mentions are coreferent is missing, the choice which mention to annotate with a remote edge is underdetermined.
This leads to multiple conceptually equivalent choices that are formally distinct, opening the way for spurious disagreements.
For example, the implicit advice-giver in \cref{fig:threesents} could be marked equally well with a remote edge to \textit{anyone else} instead of \textit{you}, resulting in a structurally diverging graph (taking the presented analysis as the reference).\footnote{While additional, more restrictive guidelines could to some extent curb such confusion (e.g., by specifying that the closest appropriate mention to the left should always be chosen as the remote target), this would require the foundational layer annotators to be confident in the notion of coreference to determine which mentions are ``appropriate'', eliminating the modularity and intuitiveness we desire.}
And third, many other implicit relations relevant to coreference (e.g., implied common sense part/whole relations, via \emph{bridging}) are not exposed in the foundational layer of UCCA.
A layer that annotates identity coreference could be extended with 
such additional information in the future.


\section{The UCoref Layer}\label{sec:desc}\label{sec:identif}

The underlying hypothesis of this work is that the spans of words that form referring expressions, i.e., evoke or point back to entities and events in the world, are also grouped as semantic units in the foundational layer of UCCA.
This assumption is motivated by the fundamental principles of UCCA as a neo-Davidsonian theory:
The basic elements of a discourse are descriptions of scenes ($\approx$ events), and their basic elements are participants ($\approx$ entities).
We can thus automatically identify scene and participant units as referring.
With this high-preci\-sion preprocessing and a small set of simple guidelines for identifying other UCCA units as referring, the process of \textit{mention identification} in UCoref is very efficient.
\Cref{fig:ref-layer} illustrates how UCoref interacts with the foundational layer.
Four referents and six mentions (two singletons) 
are identified based on the criteria below.

\paragraph*{Scene and Participant units.}
The vast majority of referent mentions can be identified by two simple rules:
\begin{inparaenum}[1)]
    \item All \textbf{scene} units are considered mentions as they constitute descriptions of actions, movements, or states as defined in the foundational layer guidelines.
    \item Similarly, all \textbf{Participant} units are considered mentions as they describe entities that are contributing to or affected by a scene/event (including locations and other scenes/events).
\end{inparaenum}

Special attention should be paid to relational nouns like \textit{teacher} or \textit{friend} that both refer to an entity and evoke a process or state in which the entity generally or habitually participates.\footnote{A teacher is a person who teaches and a friend is a person who stands in a friendship relation with another person. Cf. \citet{newell-18,meyers-04}.}
According to the UCCA guidelines, these words are analyzed internally (as both \ucca{P}\slash \ucca{S} and \ucca{A} within a nested unit over the same span), in addition to the context-dependent incoming edge from their parent.
However, the inherent scene (of teaching or friendship) is merely \emph{evoked}, but not \textit{referred to}, and it is usually invariant with respect to the explicit context it occurs in.
Moreover, treating one span of words as two mentions would pose a significant complication.
Thus, we consider these units only in their role as Participant (and not scene) mentions.

\paragraph*{Non-scene-non-participant units.}
A certain subset of the remaining unit types are considered to be mention \textit{candidates}.
This subset is comprised of the categories, \textit{Time}, \textit{Elaborator}, \textit{Relator}, \textit{Quantity}, and \textit{Adverbial}.
We give detailed guidelines for these categories, as well as for coreference markup, in the supplementary material (\cref{sec:ucoref-guidelines}).

\paragraph{Center units.}
For simplicity, a referring unit with a \emph{single} Center usually does not require its Center to be marked separately, as a unit always corefers with its Center (see \cref{sec:compare-guidel,sec:exp1} about how this relates to the min/max span distinction).

\emph{Multi}-Center units receive a different treatment:
One use of multi-Center units is coordination, where each conjunct is a Center.
Here we do want to mark up the conjuncts in addition to the whole coordination unit---provided the whole unit is referring by one of the other criteria---and assign them to separate coreference clusters.
Another class of multi-Center units, which we call \textit{relative partitive constructions}, is less straightforward to handle. Consider a phrase like \textit{the top of the mountain}.
The intuition given in the UCCA guidelines is that while the phrase is syntactically and, to some extent, semantically headed by \textit{top}, it can only be fully understood in relation to \textit{mountain}; thus, both words should be Centers.
This construction is clearly less symmetric than coordination, but at this point we do not have a reliable way of formally distinguishing the two in preprocessing, purely based on the UCCA structure and categories.
Thus, multi-Center units deserve a more nuanced manual UCoref analysis in future work;
however, for the sake of consistency and simplicity, we treat all multi-Center units in the same way as we treat coordinations in our pilot annotation (\cref{sec:anno}).

\paragraph{Implicit units.}
Implicit units may be identified as mentions and linked to coreferring expressions just like any other unit, as long as they meet the criteria outlined above. 

\finalversion{
Implicit units are a major challenge for automatic UCCA parsing and evaluation as they do not correspond to any actual tokens in the text, which limits the criteria for uniquely identifying and computing expressive features for them.
Under the proposed framework, they can be identified by their coreference clusters and features for parsing may be computed based on coreferring words.\nss{If we want to talk about UCCA parsing we should cite those papers. But I'm not sure this paragraph is crucial.}
}


\section{Comparison with Other Schemes}\label{sec:compare}\label{sec:compare-guidel}

The task of coreference resolution is far from trivial and has been approached from many different angles.
Below we give a detailed analysis of the theoretical differences between three particular frameworks: OntoNotes \citep{ontonotes}, Richer Event Description \citep[RED;][]{ogorman-16}, and the Georgetown University Multilayer corpus \citep[GUM;][]{zeldes-17}.

\paragraph*{Singletons and events.}
RED and UCoref annotate all nominal entity, nominal event, and verbal event mentions, including singletons.\footnote{For event coreference specifically, see also EventCorefBank \citep[ECB;][]{bejan-10} and the TAC-KBP Event Track \citep{mitamura-15}, which uses the ACE 2005 dataset \citep[\href{https://catalog.ldc.upenn.edu/LDC2006T06}{LDC2006T06};][]{doddington-04}.}
OntoNotes does not include singleton mentions in the coreference layer.\footnote{A separate layer records all \emph{named} entities, however, and non-coreferent ones can be considered singleton mentions.}
Further, only those verbal mentions that are coreferent with a nominal are included.
GUM includes all nominal mentions, including singletons and nominal event mentions, and follows the OntoNotes guidelines for verbal mentions.

\paragraph*{Syntactic vs.~semantic criteria.}
GUM and OntoNotes, despite not being \emph{anchored} in syntax, specify syntactic criteria for mention and coreference annotation.
The criteria in RED and UCoref, on the other hand, are fundamentally semantic.
Rough syntactic guidance is only given where appropriate and at no time is a decisive factor.

\paragraph*{Minimum and maximum spans.}
The policy on mention spans is often one of two extremes: \textit{minimum} spans (also called \textit{triggers} or \textit{nuggets}), which typically only consist of the head word or expression that sufficiently describes the type of entity or event; or \textit{maximum} spans (also called \textit{full mentions}), containing all arguments and modifiers.
GUM and OntoNotes generally apply a maximum span policy for nominal mentions, with just a few exceptions.\footnote{The GUM guidelines specify that clausal modifiers should not be included in a nominal mention.}
For verbal mentions, OntoNotes chooses minimum spans, whereas GUM annotates full clauses or sentences.
RED always uses minimum spans, except for time expressions, which follow the TIMEX3 standard \citep{pustejovsky-10}.
One of the main advantages of UCoref is that the preexisting predicate-argument and head-modifier structures of the foundational layer allow a flexible and reliable mapping between minimum and maximum span annotations.
Additionally, UCoref has `null' spans, corresponding to implicit units in UCCA.\footnote{The coreference layer of the Prague Dependency Treebank \citep{nedoluzhko-16}, quite similarly to the proposed framework, marks null-mentions arising from control verbs, reciprocals, and dual dependencies (in general, null-nodes arising from obligatory valency slot insertions into the tectogrammatical layer)---the syntactic equivalents of implicit units and remote edges in UCCA. Further, in case the mention is a root of a nontrivial subtree, it is underspecified whether the mention spans only the root, the whole subtree or some part of it.}

\paragraph*{Predication.}

OntoNotes does not assert a coreference relation between copular arguments.\footnote{Neither do \citet[in the ARRAU corpus;][]{poesio-08}.}
RED distinguishes several relation types depending on the ``predicativeness'' of the expression and in particular asserts a set-membership (i.e., non-identity) relation when the second argument is indefinite.
In GUM, relation types are assigned based on different criteria,\footnote{In particular, the notion of \textit{bridging} is interpreted differently between GUM and RED: GUM reserves it for entities that are expected (from world knowledge) to stand in some relationship (e.g., part/whole) with each other, which is reflected in a definite initial mention of the `bridging target' (\textit{\underline{My car} is broken; it's \textbf{the motor}}). RED uses it for copular predications involving relational/occupational nouns like \textit{\textbf{John} is \textbf{a/the killer}}, which are simple `coref' (or `ana'/`cata', if one mention is a pronoun) relations in GUM.
We consider neither of these definitions in this work (see \cref{sec:ucoref-coref-guidelines}).} %
and, depending on the polarity and modality of the copula, its arguments may be marked as coreferring mentions, even if they are indefinite.\footnote{See also \citet{chinchor-98}.}
A slightly different distinction is made in UCoref, where, thanks to the foundational layer, evokers of set-membership and attributive relations are marked as stative scenes in which the modified entity participates.
Definite identity is handled in the same way as in RED, as well as relational nouns except for the special case of generic mentions (\cref{sec:ucoref-coref-guidelines}).

\paragraph*{Apposition.}

In RED and OntoNotes, punctuation is considered a strict criterion for marking appositives, while GUM relies solely on syntactic completeness.
In OntoNotes and GUM, ages specified after a person's name are considered separate appositional mentions, coreferring with the name mention they modify.
UCoref takes advantage of UCCA's semantic Center-Elaborator structure, abstracting away from superficial markers like punctuation which may not be available in all genres and languages (details in \cref{sec:ucoref-coref-guidelines}).

\paragraph*{Prepositions.}
Whereas OntoNotes and GUM stick to the syntactic notion of NPs, UCoref includes prepositions and case markers within mentions.
This does not have a major effect on coreference, but contributes to consistency between languages that vary in the grammaticalization of their case marking.

\paragraph*{Coordination.}
Our treatment of coordinate entity mentions is adopted and expanded from the GUM guidelines, where the span containing the full coordination is only marked up if it is antecedent to a plural pronominal mention.
OntoNotes does not specify how coordinations in particular should be handled; while the guidelines state that out of \emph{head-sharing} (i.e., elliptic) mentions only the largest one should be picked, we assume that coordinations of multiple \emph{explicitly headed} phrases are not targeted as mentions in addition to the conjuncts.
The minimum span approach of RED precludes marking full coordinations in addition to conjuncts.

\paragraph*{Summary.}
That OntoNotes does not annotate singleton mentions makes it the most restrictive out of the compared frameworks.
Despite its emphasis on syntax, GUM is closer to our framework as it includes singletons and marks full spans for non-singleton events;
the marking of bridging relations, directed coreference links, and information status present in GUM is beyond our scope here.
RED is conceptually closest to UCoref in marking all entity, time, and event mentions, except for the difference in span boundaries.
This can largely be resolved as we will show in \cref{sec:exp1}.

\section{Pilot Annotation}\label{sec:pilot}

In order to evaluate the accessibility of the annotation guidelines given above and in \cref{sec:ucoref-guidelines}, and facilitate empirical comparison with other schemes, we conducted a pilot annotation study. 
\label{sec:anno}
We annotated a small English dataset consisting of subsets of the OntoNotes (\href{https://catalog.ldc.upenn.edu/LDC2013T19}{LDC2013T19}), RED (\href{https://catalog.ldc.upenn.edu/LDC2016T23}{LDC2016T23}), and GUM\footnote{\url{https://github.com/amir-zeldes/gum}} corpora with the UCCA foundational and coreference layers.%
\footnote{Since the RED documents are not tokenized (character spans are used for mention identification), we preprocessed them with the PTB tokenizer and the Punkt sentence splitter using Python NLTK.}

The OntoNotes documents are taken from blog posts, 
the GUM documents are WikiHow instructional guides, and the RED documents are online forum discussions.
Because all annotations were done by a single annotator each and not reviewed, our results are to be understood as a proof of concept; 
measuring interannotator agreement will be necessary in the future to gauge the difficulty of the task and quality of guidelines\slash data.
\finalversion{UCCA foundational layer annotation on the OntoNotes and RED data was done by the first author and the GUM documents were annotated by multiple semi-expert graduate students.
The UCoref annotations for all documents were done by the first author, while iteratively developing the annotation guidelines.}

\begin{table}[t]\centering\small
         \begin{tabular}{|l|c|c|c|}
\hline
& GUM & OntoNotes & RED \\\hline
sentences & \phantom{00}70 & \phantom{0}17 & \phantom{0}24  \\
tokens & 1180 & 303 & 302 \\
$\hookrightarrow$ non-punct & 1030 & 261 & 274 \\\hline
UCCA units  & 1436  & 336  & 379 \\
$\hookrightarrow$ candidates  & \phantom{0}911  & 195 & 186 \\
\hline
    \end{tabular}
    \caption{Overview of our pilot dataset. \textit{Candidates} refers to the UCCA units that are filtered by category for mention candidacy before manual annotation.}
    \label{tab:sentences}
\end{table}

\begin{table*}[ht]\centering\small\setlength{\tabcolsep}{4pt}
         \begin{tabular}{|l|c|c||c|c||c|c||l|c|c||c|c||c|c|}
\multicolumn{1}{c}{} & \multicolumn{2}{c}{\textbf{WikiHow}} & \multicolumn{2}{c}{\textbf{Blog}} & \multicolumn{2}{c}{\textbf{Forum}} & \multicolumn{1}{c}{} & \multicolumn{2}{c}{\textbf{WikiHow}} & \multicolumn{2}{c}{\textbf{Blog}} & \multicolumn{2}{c}{\textbf{Forum}} \\
\cline{2-7}\cline{9-14}
\multicolumn{1}{c|}{} & GUM & UCR & ONT & UCR & RED & UCR & \multicolumn{1}{c|}{}  & GUM & UCR & ONT & UCR & RED & UCR \\\hline
\textbf{mentions} & 288 & \phantom{}466 & 40 & 128 & 120 & 117 & \textbf{referents} & 155 & \phantom{}291 & 20 & \phantom{}96 & \phantom{}82 & \phantom{}78 \\\hline
$\hookrightarrow$ event & 158 & \phantom{}208 & -- & \phantom{0}47 & \phantom{0}70 & \phantom{0}54 & $\hookrightarrow$ event & 108 & \phantom{}180 & -- & \phantom{}43 & \phantom{}58 & \phantom{}47 \\
$\hookrightarrow$ entity/\ucca{A} & 127 & \phantom{}215 & -- & \phantom{0}66 & \phantom{0}47 & \phantom{0}58 & $\hookrightarrow$ entity & \phantom{0}47 & \phantom{}108 & -- & \phantom{}46 & \phantom{}21 & \phantom{}27  \\
$\hookrightarrow$ other & \phantom{00}3 & \phantom{0}43 & -- & \phantom{0}14 & \phantom{00}3 & \phantom{00}5 & $\hookrightarrow$ time & \phantom{00}0 & \phantom{00}3 & -- & \phantom{0}7 & \phantom{0}3 & \phantom{0}4 \\\hline
$\hookrightarrow$ NE & -- & -- & \phantom{}10 & -- & -- & -- & $\hookrightarrow$ non-singleton & \phantom{0}46 & \phantom{0}36 & 10 & 13 & \phantom{0}9 & 18 \\
$\hookrightarrow$ IMP & -- & \phantom{0}26 & -- & \phantom{00}6 & -- & \phantom{00}4 & \hspace{1em}$\hookrightarrow$ IMP & -- & \phantom{0}26 & -- & \phantom{0}1 & -- & \phantom{0}4 \\
$\hookrightarrow$ remote & -- & \phantom{0}10 & -- & \phantom{00}3 & -- & \phantom{00}1 & \hspace{1em}$\hookrightarrow$ remote & -- & \phantom{00}7 & -- & \phantom{0}2 & -- & \phantom{0}1 \\
\hline
\end{tabular} %
\caption{\label{tab:stats}Distribution of mentions and referents in the datasets. 
\textbf{Mentions}: Under \textit{event}, we count UCoref (UCR) scenes, GUM mentions of the types `event' or `abstract', and RED EVENTs;
under \textit{entity}, we count UCR \ucca{A}'s, GUM `person', `object', `place', and `substance' mentions, and RED ENTITYs.
\textit{NE} stands for OntoNotes (ONT) named entities and IMP and remote for implicit and remote UCR units.
A coreference cluster (\textbf{referent}) is classified as an \emph{event} referent if there is at least one event mention of that referent, as a \emph{time} referent if there is at least one UCR \ucca{T} / GUM `time' / RED TIMEX3 mention of that referent, and as an \emph{entity} referent otherwise; we also report how many of the IMP and remote units are part of non-singleton referents.
}
\end{table*}

\Cref{tab:sentences} shows the distribution of tokens and UCCA foundational units, and \cref{tab:stats} compares the distribution of UCoref units with the respective ``native'' annotation schema for each corpus.
We can see that about one third of all UCCA units are identified as mentions, in all corpora.
The automatic candidate filtering based on UCCA categories simplifies this process for the annotator by removing about one third to one half of units.
There are similar amounts of scene and Participant units (both of which are always mentions), but it is important to note that Participant units can also refer to events.
We can see this reflected by the majority of referent units being event referents.
We can also see that most of the referents in GUM, RED, and UCoref are in fact singletons, and the number of non-singleton referents is quite similar between each scheme and UCoref.
Most implicit units and targets of remote edges are part of a non-singleton coreference cluster, which confirms the issue of spurious ambiguity we pointed out in \cref{sec:ucca-motiv}.

\subsection{Recovering Existing Schemes}\label{sec:exp1}

Next we examine the differences in gold annotations between our proposed schema and existing schemas and how we can (re)cover annotations in established schemas from our new schema.
We can interpret this experiment as asking: If we had a perfect system for UCoref, could we use that to predict GUM/OntoNotes/RED-style coreference?
And vice versa, if we had an oracle in one of those schemes, and possibly oracle UCoref mentions, how closely could we convert to UCoref?\footnote{See also \citet{zeldes-16-xrenner}, who base a full coreference resolution system on this idea.}

\paragraph*{Exact mention matches.}

A na\"{i}ve approach would be to look at the token spans covered by all mentions and reference clusters and count how often we can find an exact match between UCoref and one of the existing schemes.

In UCoref, we use maximum spans by default, but thanks to the nature of the UCCA foundational layer, minimum spans can easily be recovered from Centers and scene-evokers.
For schemas with a \emph{minimum span} approach, we can switch to a minimum span approach in UCoref by choosing the head unit of each maximum span unit as its representative mention.
This works well between UCoref and RED as they have similar policies for determining semantic heads, which is crucial for, e.g., light verb constructions.
This would be problematic, however, when comparing to a minimum span schema that uses syntactic heads.
For schemas with a \emph{non-minimum span} approach, we keep only the maximum span units from UCoref and discard any heads that have been marked up representatively for their parent (e.g., as remote targets).

\begin{table*}[ht]\small\centering\setlength{\tabcolsep}{3.5pt}
    \centering
    \begin{tabular}{|r|ccc|ccc|ccc|r|ccc|ccc|ccc|}
    \hline
          \multicolumn{10}{|c|}{\textbf{mentions}} & \multicolumn{10}{c|}{\textbf{referents}} \\\hline
          & \multicolumn{3}{c|}{GUM} & \multicolumn{3}{c|}{OntoNotes} & \multicolumn{3}{c|}{RED} & & \multicolumn{3}{c|}{GUM} & \multicolumn{3}{c|}{OntoNotes} & \multicolumn{3}{c|}{RED} \\
          & P & R & F & P & R & F & P & R & F & & P & R & F & P & R & F & P & R & F \\\hline\hline
\multirow{2}{*}{$=$} & \multirow{2}{*}{41.7} & \multirow{2}{*}{60.6} & \multirow{2}{*}{49.4} & \multirow{2}{*}{21.3} & \multirow{2}{*}{\phantom{0}67.5} & \multirow{2}{*}{32.3} & \multirow{2}{*}{\textit{79.5}} & \multirow{2}{*}{\textit{77.5}} & \multirow{2}{*}{\textit{78.5}} & $=$ & 19.3 & 24.4 & 21.6 & \phantom{0}3.1 & 15.2 & \phantom{0}5.2 & \textit{52.6} & \textit{50.0} & \textit{51.3} \\
& & & & & & & & & & $\approx$ & 31.6 & 40.0 & 35.3 & \phantom{0}9.4 & 45.0 & 15.5 & \textit{66.7} & \textit{63.4} & \textit{65.0} \\\hline
\multirow{2}{*}{$\approx$} & \multirow{2}{*}{67.0} & \multirow{2}{*}{97.2} & \multirow{2}{*}{79.3} & \multirow{2}{*}{31.5} & \multirow{2}{*}{100.0} & \multirow{2}{*}{47.9} & \multirow{2}{*}{\textit{88.9}} & \multirow{2}{*}{\textit{86.7}} & \multirow{2}{*}{\textit{87.8}} & $=$ & 43.9 & 55.6 & 49.0 & \phantom{0}5.2 & 25.0 & \phantom{0}8.6 & \textit{66.7} & \textit{63.4} & \textit{65.0} \\
& & & & & & & & & & $\approx$ & 59.6 & 75.6 & 66.7 & 10.4 & 50.0 & 17.2 & \textit{80.8} & \textit{76.8} & \textit{78.8} \\\hline
    \end{tabular}
    \caption{Exact ($=$) and fuzzy ($\approx$) referent matches based on exact and aligned mentions between UCoref and GUM, OntoNotes, and RED.
    Precision (P) and recall (R) are measured treating gold UCoref annotation as the prediction and gold annotation in each respective existing framework as the reference. Italics indicate minimum UCoref spans are used. 
    Implicit UCoref units are excluded from this evaluation, and children of remote edges are only counted once (for their primary edge).
    }
    \label{tab:match}
\end{table*}

\paragraph*{Fuzzy mention matches.}

Because our theoretical comparison in \cref{sec:compare-guidel} exposed systematically diverging definitions of what to include in a mention span, we also apply an evaluation that abstracts away from some of these differences.
We greedily identify one-to-one alignments for maximally overlapping mentions, as measured by the Dice coefficient:%

\[
m^*_A, m^*_B \leftarrow \argmax_{m_A\in (A \setminus L_A),\, m_B\in (B \setminus L_B)}{\frac{|m_A \cap m_B|}{|m_A| + |m_B|}} 
\]
where $L_A$ ($L_B$) records the mentions from annotation $A$ ($B$) aligned thus far, and stopping when this score falls below a threshold $\mu$.
$\mu$ is a hyperparameter controlling how much overlap is required: $\mu=1$ corresponds to exact matches only, while $\mu=0$ includes all overlapping mention pairs as candidates (we report fuzzy match results for $\mu=0$). 
Once a mention is aligned it is removed from consideration for future alignments.

We align referents by the same procedure. Results are reported in \cref{tab:match}.

\subsection{Findings}

We can see in \cref{tab:match} that UCoref generally covers between 60\% and 80\% of exact \emph{mentions} in existing schemes (`R' columns), however, the amount of UCoref units that are present in other schemes varies greatly, between 21.3\% (OntoNotes) and 79.5\% (RED; `P' columns).
This is generally expected based on our theoretical analysis in \cref{sec:compare-guidel}.
Fuzzy match has a great effect on the maximum span schemes in GUM and OntoNotes, resulting in up to 100\% of mentions being aligned, and a lesser, but still positive effect on RED.\footnote{Note, though, that this evaluation only shows us \emph{if} we can find a fuzzy alignment, not whether the aligned spans are actually equivalent.
As purely span-based alignment is prone to errors, a future extension to the algorithm should take information about (ideally semantic) heads into account.}
We observe a similar trend for \emph{referent} matches, which follows partly from the mismatch in mention annotation, and partly from diverging policies in marking coreference relations, as discussed above. Whether or not singleton event and\slash or entity referents are annotated has a major impact here.
Below we give examples for sources of non-exact mention matches that can be resolved using fuzzy alignment.

\subparagraph*{GUM and OntoNotes.}

A phenomenon that is trivially resolvable using fuzzy alignments is punctuation, which is excluded from all UCoref units, but included in GUM and OntoNotes.
Another group of mentions recovered are prepositional phrases, where UCoref includes prepositions (\textit{to them}, \textit{since the end of 2005}), and GUM and OntoNotes do not (\textit{them}, \textit{the end of 2005}).
As mentioned in \cref{sec:compare-guidel}, GUM deviates from its maximum span policy for clausal  modifiers of noun phrases, which are stripped off from the mention.
Noun phrases modified in this way can be fuzzily aligned with the maximum spans in UCoref, even if the modifier is very long:
\textit{people who are stuck on themselves intolerant of people different from them rude or downright arrogant} (UCoref) gets aligned with \textit{people} (GUM).

\subparagraph*{RED.}

Almost 80\% of both RED and UCoref mentions match exactly, but there are some cases of divergence:
1) One subset of these are time expressions like \textit{this morning}, where, as pointed out above, RED marks maximum spans.
However, in UCoref these are internally analyzable---thus their Center will be extracted for minimum spans (here, \textit{morning}).
On the other hand, idiomatic multiword expressions (MWEs) such as verb-particle constructions (e.g., \textit{pass away} `die') are treated as unanalyzable in UCCA, but only the syntactic head (\textit{pass}) is included in RED.
2) Also interesting are predicative prepositions and adverbials in copular or expletive constructions:
\textit{there will be lots of good dr.s and nurses around}.
Here, UCoref chooses \textit{around} as the (stative) scene evoker (and would mark the prepositional object as a participant, if it is explicit), while RED chooses the copula \textit{be}.
3) UCCA treats some verbs as modifiers rather than predicates themselves: e.g., \textit{stopped} in \textit{i m \emph{[sic]} stopped feeling her move} and \textit{it seemed} in \textit{it seemed tom \emph{[sic]} take forever}. 
The former, as an aspectual secondary verb, is labeled Adverbial~(\ucca{D}); 
the latter, which injects the perspective of the speaker, is labeled Ground~(\ucca{G}).
Since we do not generally consider these categories referring, these are not annotated as mentions in UCoref, though they are in RED.

\subsection{Discussion}

For the non-minimum span schemas GUM and OntoNotes, we can use a fuzzy mention alignment based on token overlap to find many pairs which aim to capture the same mention, only under different annotation conventions.
RED is most similar to UCoref in defining what counts as a mention, though our corpus analysis showed that the notion of \emph{semantic heads} is interpreted differently for certain constructions, where UCCA is more liberal about treating verbs as modifiers rather than heads.
While counting fuzzy matches allows us to recover partially overlapping spans (time expressions, verbal MWEs), other phenomena (adverbial copula constructions, secondary verbs) have inconsistent policies between the two schemes that require more elaborate methods to align.
We can thus, to some extent, use UCoref to predict RED-style annotations, with the additional gain of flexible minimum/maximum spans and cross-sentence predicate-argument structure for a whole document.
Furthermore, we see that UCoref subsumes all OntoNotes mentions and nearly all GUM mentions and is able to reconstruct coreference clusters in GUM with high recall.

\section{Conclusion}

We have defined and piloted a new, modular approach to coreference annotation based on the semantic foundational layer provided by UCCA. 
An oracle experiment shows high recall with respect to three existing schemes, as well as high precision with respect to the most similar of the three.
\shortversion{We will release our annotations upon publication.}
We have released our annotations to fuel future investigations.

\section*{Acknowledgments}
We would like to thank Amir Zeldes and two anonymous reviewers for their many helpful comments, corrections, and pointers to relevant literature.
This research was supported in part by NSF award IIS-1812778 and grant 2016375 from the United States--Israel Binational Science Foundation (BSF), Jerusalem, Israel.

\bibliography{ms}
\bibliographystyle{acl_natbib}

\clearpage

\appendix

\section{Detailed Guidelines}\label{sec:ucoref-guidelines}

\subsection{Identifying Mentions}\label{sec:ucoref-mention-guidelines}

\paragraph*{Non-scene-non-participant units.}
A certain subset of the remaining unit types are considered to be mention \textit{candidates}.
This subset is comprised of the categories, \textit{Time}, \textit{Elaborator}, \textit{Relator}, \textit{Quantity}, and \textit{Adverbial}.

\subparagraph{Time (\ucca{T})}
Absolute or relative time expressions like \textit{on May 15, 1990}, \textit{now}, or \textit{in the past}, which are marked Time (\ucca{T}) in UCCA, are considered mentions.
However, frequencies and durations, which are also \ucca{T} units in UCCA, are discarded.
In order to reliably distinguish these different kinds of time expressions from each other, they have to be identified manually.

\subparagraph{Elaborator (\ucca{E})}
Elaborators modifying the Center (\ucca{C}) of a non-scene unit are considered mentions if they themselves describe a scene or entity.
This is the case, for example, with (relative) clauses and (prepositional) phrases describing the Center's relation with another entity as in

\begin{center}
\textit{\underline{[ the book$_C$ \underline{[about the dog$_C$ ]$_E$} ]},}\\
\end{center}

\noindent as well as contingent attributive modifiers, which are stative scenes in UCCA, like \textit{old} in

\begin{center}
\textit{\underline{[ the \underline{[ old$_S$ (book)$_A$ ]$_E$} book$_C$ ]}}.\\
\end{center}

By contrast, Elaborator units that do not evoke a person, thing, abstract object, or scene are not considered referring, as in

\begin{center}
\textit{\underline{[ medical$_E$ school$_C$ ]}},\\
\end{center}

\noindent where \textit{medical} is an inherent property and thus non-referring.

In English, this often corresponds to units whose Center is an adjective, adverb, or determiner.\footnote{According to the UCCA v1 guidelines, articles are to be annotated as Elaborators. In the v2 guidelines, the default category for articles has changed to Function.}
Bear in mind, however, that these syntactic criteria are language-specific and should only be taken as rough guidance, rather than absolute rules.
Thus, referring non-scene Elaborators should be identified manually.
By contrast, \ucca{E}-scenes will be identified as mentions automatically, by the scene unit criterion.

\subparagraph{Relator (\ucca{R})}
Relators should be marked as mentions if and only if they constitute an anaphoric (or cataphoric) reference \emph{in addition to} their relating function.

As an illustration what we mean by that, consider the two occurrences of \textbf{that} in the following example, which are both Relators in UCCA:

\begin{center}
\textit{I didn't like \textbf{that}$_1$ he said the things \textbf{that}$_2$ he said.}
\end{center}

Here, \textit{that}$_2$ is an anaphoric reference to \textit{things}, whereas \textit{that}$_1$ is purely functional and thus should not be identified as referring.
In English, this corresponds to the syntactic category of relative pronouns.

Most \ucca{R} units, however, are non-referring expressions like prepositions, so identification of the few referring instances of Relators has to be done manually.

\subparagraph{Quantity (\ucca{Q})}
Partitive constructions like \textit{one of the 5 books} contain mentions of two distinct referents: \textit{the 5 books} and \textit{one of the 5 books}.
According to the v2 UCCA guidelines, these expressions should be annotated as an Elaborator-Center structure with a remote edge:

\begin{center}
\textit{\underline{[[ one$_Q$ (books)$_C$ ]$_C$ \underline{[ of$_R$ the$_F$ 5$_Q$ books$_C$ ]$_E$} ]$_X$}}\\
\end{center}

Such an annotation will result in correct identification of the two mentions based on the guidelines given so far (by choosing the \ucca{E} unit and the whole X\footnote{We use the placeholder X here as the actual category depends on the context (i.e., sibling and parent units) in which a unit is embedded.} unit), without the need to identify the Quantifier (\ucca{Q}) unit \textit{one$_Q$}.
However, in foundational layer annotations made based on the UCCA v1 guidelines the same phrase would receive a flat structure (cf.~discussion of Centers above):

\begin{center}
\textit{\underline{[ one$_Q$ of$_R$ the$_F$ \underline{5$_Q$} books$_C$ ]$_X$}}\\
\end{center}

In this case, we choose the whole X unit as a mention of the one book (respecting semantics rather than morphology), and the \ucca{Q} unit \textit{5} as a mention of the five books.

\subparagraph{Adverbial (\ucca{D})}
While most Adverbial units (\ucca{D}) are by default not considered to be referring (they describe secondary relations over events), in some cases \ucca{D} units can be identified as mentions (also see \textbf{coordinated mentions} in \cref{sec:ucoref-coref-guidelines}).

One such phenomenon are prepositional phrases like \textit{for another reason} and \textit{in the majority of cases} are annotated as \ucca{D} in the corpus, as they modify scenes, not entities.

Another class of Adverbial units that may be identified as referring are the so-called secondary verbs like \textit{help}, \textit{want} and \textit{offer}, which, according to the UCCA guidelines, modify scenes evoked by primary verbs, but do not themselves denote scenes.
However, the relations described by them can sometimes be coreferring antecedents independently from the main scene:

\begin{center}
\textit{\underline{[ I$_A$ {lost$_P$} [ 10 lbs ]$_A$ . ]}$_j$}\\
\textit{[ I$_A$ was$_F$ extremely$_D$ happy$_S$ \underline{[ about$_R$ that$_C$ ]$_A$}$_j$ ] .}\\

vs.

\textit{[ She$_A$ \underline{helped$_D$}$_i$ me$_A$ {lose$_P$}$_j$ [ 10 lbs ]$_A$ . ]}\\
\textit{[ I$_A$ really$_D$ appreciated$_P$ \underline{that$_A$}$_i$ ] .}\\
\end{center}

In both examples, \textit{losing weight} is the main scene according to UCCA, however, in the second example the object of appreciation is \textit{helping}.
Thus, we do mark secondary verbs as mentions, but \textit{only if} they are referred back to in the way demonstrated above.

\subsection{Resolving Coreference}\label{sec:ucoref-coref-guidelines}

\paragraph*{Appositives.}
Appositives and titles cooccurring with (named) entity mentions are annotated as Elaborators in UCCA and thus automatically included in the entity mention they modify.
They should be marked as separate mentions, coreferring with the modified unit.

If a title or occupational noun occurs by itself or as a copular argument, we treat it as a relational noun as described in the next paragraph.

\paragraph*{Extensional vs.~intensional readings.}
A coordinated mention of a group of individuals, such as \textit{John, Paul, and Mary} evokes a referent that is distinct from the possibly already evoked referents of \textit{John}, \textit{Paul}, and \textit{Mary}, respectively.

Relational nouns (e.g., ``the president [of Y]''), which are instantiated by a specific individual or a fixed-size set of individuals (e.g., ``[X's] parents'') at any given point in time, should usually be marked as coreferring with their instances, as inferred from context.
This corresponds to an \textit{extensional} (or set-theoretical) notion of reference: a distinct referent is identified by the individuals in which this concept manifests (extension).

Only in clearly generic statements like

\begin{center}
\textit{The president's power is limited by the constitution.}\\[5px]
\textit{You should always do what your parents tell you to.}
\end{center}

\noindent should they evoke separate referents from any specific presidents or parents also mentioned in the same discourse.
This corresponds to an \textit{intensional} (or indirect) notion of reference: a distinct referent is identified by its general idea or concept (intension), rather than its instances.

Mentions of group-like entities with undetermined size, such as \textit{the committee} or \textit{all committee members}, should always be analyzed intensionally, evoking a referent separate from the possibly mentioned referents for the individuals comprising it.\footnote{But singular and plural mentions of the same group can corefer \citep{zeldes-18}.}

\paragraph*{Negated scenes.}
Mentions of scenes are referring and should be marked as coreferring with other mentions of the same scene (same process or state and same participants), regardless of whether or not that scene really took place or is hypothetical:

\begin{center}
\textit{I hoped \underline{she liked the pizza}$_i$ and was relieved when I learned \underline{she did}$_i$.}\\
\end{center}

When both a scene and its negation are mentioned, these mentions should evoke separate referents:

\begin{center}
\textit{I hoped \underline{she liked the pizza}$_i$ and was surprised when I learned \underline{she didn't}$_j$.}\\
\end{center}

\paragraph*{Coordinated mentions.}
When entities or events are described in conjunction, they evoke a separate referent for each of the conjuncts, and a third one for the set comprising them. 
The whole coordination can be explicitly referred to with another (pronominal) mention:

\begin{center}
\textit{It is likely \underline{[ that \underline{[ the shock will dislocate ]}$_i$ \textbf{and}} \underline{\underline{[ (the shock) break both your arms ]}$_j$ ]}$_k$; nevertheless \underline{this}$_k$ is a small price to pay for your life.}\\
\end{center}

\begin{center}
\textit{I want \underline{[ \underline{Ivy}$_i$ \textbf{and} \underline{William}$_j$ ]}$_k$ on my debate team, because \underline{[ both of them ]}$_k$ are great.}\\
\end{center}

However, if the mentions are presented in \textit{dis}junction, no separate mention for the full disjunction should be marked.
If an anaphoric pronoun occurs, there are several options.

For events, a secondary relation (marked \ucca{D} in UCCA) that holds for both conjuncts, if available, should be marked instead:

\begin{center}
\textit{It is \underline{likely}$_k$ [ that \underline{[ the shock will dislocate ]}$_i$ \textbf{or} \underline{[ (the shock) break both your arms ]}$_j$ ] ; nevertheless \underline{this}$_{k}$ is a small price to pay for your life.}\\
\end{center}

If such a unit is not available, and for entities, no coreference relation exists:

\begin{center}
\textit{I want [ \underline{Ivy}$_i$ \textbf{or} \underline{William}$_j$ ] on my debate team, because \underline{[ both of them ]}$_k$ are great.}\\
\end{center}

\paragraph*{Remote edges.}
Different types of remote edges call for different coreference annotations.
\emph{Non}-head (i.e., non-Center, -State, or -Process) remote edges indicate that the same entity/scene modifies or participates in two (potentially also coreferent) unit mentions, namely its primary parent (or the primary parent of the unit it heads) and its remote parent.
This corresponds to zero anaphora, or a ``core'' element that is implicit in one context and explicit in another.
\emph{Head} remote edges, however, merely indicate the \emph{category} of entity/event that is shared between a full and an elliptic or anaphoric mention \citep[``sense anaphora'';][]{recasens-16}.
E.g., \textit{books} in ``two of the 5 books'' is category-shared between \textit{5 books} and \textit{two (books)}, which are separate non-coreferent mentions.
Whether the primary and remote parent are coreferent or not is contingent on context.

\end{document}